\documentclass[letterpaper, 10 pt, conference]{IEEEconf}      
 \IEEEoverridecommandlockouts                              
\usepackage{graphicx}
\usepackage{float}
\graphicspath{ {images/} }
\usepackage{caption}
\usepackage{subcaption}
\usepackage{amsmath}
\usepackage{cite}
\usepackage{algorithm}
\usepackage{algpseudocode}
\usepackage{url}
\usepackage{hyperref}
\usepackage{xcolor}
\usepackage{comment}
\usepackage{array}

\newcommand{\orcid}[1]{\href{https://orcid.org/#1}{\includegraphics[width=10pt]{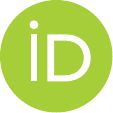}}}

\begin{document}

\title{\LARGE \bf Force Aware Branch Manipulation To Assist Agricultural Tasks \\
\author{ Madhav Rijal$^{1}$ \orcid{0000-0001-9400-0877}, Rashik Shrestha \orcid{0009-0007-7939-2623}, Trevor Smith\orcid{0000-0002-8921-9281}, Yu Gu \orcid{0000-0003-3165-3269}\vspace{2mm}\\
\textit{West Virginia University}\\
Morgantown, USA}
\thanks{The authors are with the Department of Mechanical and Aerospace Engineering, West Virginia University, Morgantown, WV 26505, USA}
\thanks{$^{1}$ Madhav Rijal is the corresponding author 
{\tt\small mr00059@mix.wvu.edu}}%

\thanks{This study was supported in part by USDA NIFA Award 2022-67021-36124 "Collaborative Research: NRI: StickBug - an Effective Co-Robot for Precision Pollination", Statler fellowship Award and the National Science Foundation Graduate Research Fellowship Award \#2136524.}

}

\maketitle
\begin{abstract}

This study presents a methodology to safely manipulate branches to aid various agricultural tasks. Humans in a real agricultural environment often manipulate branches to perform agricultural tasks effectively, but current agricultural robots lack this capability. This proposed strategy to manipulate branches can aid in different precision agriculture tasks, such as fruit picking in dense foliage, pollinating flowers under occlusion, and moving overhanging vines and branches for navigation. The proposed method modifies RRT* to plan a path that satisfies the branch geometric constraints and obeys branch deformable characteristics. Re-planning is done to obtain a path that helps the robot exert force within a desired range so that branches are not damaged during manipulation. Experimentally, this method achieved a success rate of 78\% across 50 trials, successfully moving a branch from different starting points to a target region. 
 
 
\end{abstract}

\section{Introduction}

Occlusion handling in agricultural environments is crucial for achieving the desired efficacy of precision agricultural robotic systems. Modern agricultural robots seek to assist farmers with various tasks, such as harvesting \cite{williams2019robotic}, seed planting \cite{seeding}, spraying \cite{spraying}, pruning \cite{pruning2022silwal}, and pollinating \cite{smith2024design}. However, their inability to handle occlusions has limited their success rate to an average of 66\% (ranging from 40\% to 86\%) \cite{tang2020recognition}, even in simplified environments with improved accessibility and visibility. This has reduced the widespread adoption of robot-assisted farming technologies \cite{zhou2022intelligent}. Practices such as  new cultivation systems and genetic modification of crops for better accessibility have been adopted to reduce occlusion \cite{van2005cultivation}. However, without specialized infrastructure or crop modification, generalizable occlusion handling could be achieved by developing a robot's ability to interact with a plant's branches.

\begin{figure}
    \centering
    \includegraphics[width=0.48\textwidth]{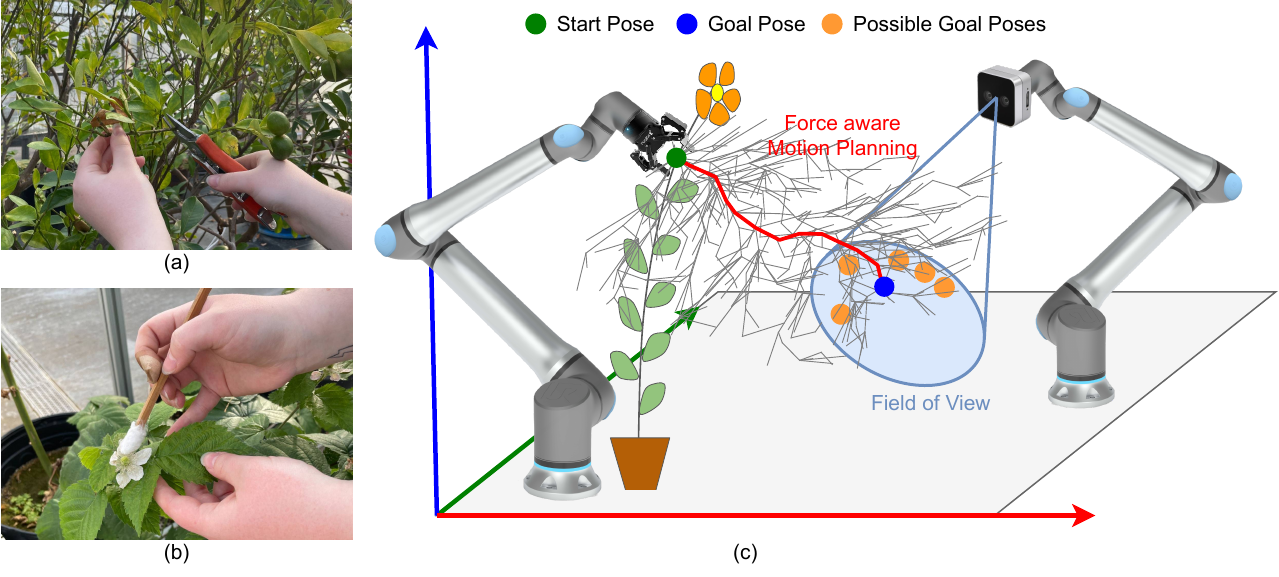}
    \caption{Humans often use one hand to grasp the branch for better accessibility, while the other hand is used to perform primary tasks like (a) branch pruning and (b) hand pollination of the flower. (c) An overview of our approach, where one robot manipulates the branch to move the flower to the field of view of another robot by planning a force-aware path.}
    \label{fig:intro}
\end{figure}

Manipulating deformable objects such as branches poses a significant challenge for robotics due to their deformability, uniqueness, and non-linear continuous dynamics \cite{deform_challenges}. These branch properties prevent conventional rigid body kinematics for motion planning, thus leading roboticists to model-based approaches. For instance, finite element methods (FEM) have been used to generate accurate models of deformable objects, but are computationally heavy\cite{fem_and_spring_2007survey}. Mass-spring-based modeling reduces computational load but still requires explicit tuning for each object \cite{manipulation_survey}. These modeling challenges create a significant gap between simulation and reality, leading to learning-based approaches to branch modeling, such as simulation driven inverse inference approach  \cite{branch_dyanmics}, which probes a physical tree to learn its dynamics.

The plant's delicacy and fixation to the ground pose significant challenges for branch manipulation. Ground fixation introduces workspace constraints that are difficult to manage, especially with high degree of freedom (DoF) manipulators that have discontinuous configuration spaces (c-spaces) \cite{berenson2011task}. Additionally, the delicate nature of branches demands high precision and minimal force to prevent damage. These factors complicate constrained motion planning \cite{berenson2011task}. To address these challenges, we propose a force-aware planning method that generates waypoints adhering to branch geometric constraints while minimizing the risk of damage.

 Overall, this work's contributions are summarized as follows:
 \begin{enumerate}

     \item  A geometric heuristic model for branches that does not require individual branch parameter tuning or probing.
     \item  A motion planning strategy for branch manipulation that adheres to workspace and branch constraints, using the geometric heuristic model to guide RRT* with online re-planning based on force feedback.
     \item  Experimental demonstration of force feedback-based motion planning protecting branches from excessive force.
     \item Software and data developed are open sourced \cite{open_source}.
     
 \end{enumerate}
 
The interactive ability to manipulate branches will enhance tasks such as harvesting, pruning, and pollination by exposing fruits, cut-points, and hidden flowers within the branch canopy. This helps overcome some bottlenecks in adopting robot-assisted agricultural technologies. The rest of the paper is outlined as follows: Section II discusses relevant works in deformable object manipulation, and their application in robot-assisted branch manipulation; Section III defines the problem statement; Section IV details the technical approach; Section V describes the experimental setup along with experimental results and discussion; Lastly, Section V concludes the work and outlines the potential future work of the study. 

\section{Related Work}
\subsection{Deformable Object Manipulation}

In robotics, flexible wires were manipulated \cite{nakagaki1997study} by modeling both plastic and elastic deformations from elasticity theory \cite{terzopoulos1987elastically}. To reduce computational costs, randomized planning algorithms were developed that minimized the elastic energy of the object \cite{lamiraux2001planning}. However, this work assumed a free-flying gripper and neglected the robot’s workspace constraints. Moreover, deformable object manipulation has been extended to two cooperative robots in \cite{saha2006motion}, where a topologically biased probabilistic roadmap is utilized for knot-tying.

Branch manipulation presents unique challenges due to their delicacy, requiring planning methods that minimize damage. This delicacy is reminiscent of surgical tasks, which utilize FEM based approaches, which with precise tuning, achieve a highly accurate model of the force interactions between microneedles and skin to design needles that reduce tissue damage \cite{kaufmann2009flexible, skin_fem}. However, FEM is computationally expensive and thus difficult for online branch manipulation.
Mass-spring models \cite{desbrun1999interactive} have been developed for animations to reduce the computational requirements of FEM but still require precise tuning of object parameters. Additionally, a model-free approach based on diminishing rigidity has been explored for manipulating deformable objects \cite{diminishing_rigidity}; however, computation still poses a significant challenge. 

Sensor feedback-based methods have also gained popularity as they avoid modeling through online feedback. For example, a kinematic controller utilizing a deformation Jacobian matrix was used to actively deform elastic objects by minimizing the error between estimated and measured deformation flows from a vision system \cite{navarro2014visual}. However, vision-based manipulation alone is often insufficient for branches due to occlusions caused by leaves. A comprehensive review has detailed various methods for modeling, shape estimation, motion planning, and control of deformable objects across applications in multiple domains\cite{review_deformation2020}.

\subsection{Robot Assisted Branch Manipulation}

Many robotic systems in the literature treat branches as obstacles to avoid, but a few have explored direct manipulation to facilitate specific tasks. Branch Pruning robots \cite{pruning2022silwal, you2022autonomouspruning} minimally interact with the branches by only making contact at the cut location with its end effector’s cutting tool. While interaction with leaves has been demonstrated in path planning by modeling them as semi-permeable objects in \cite{citrus_rrt}. Similarly, fruit interaction has been explored with motion primitives in \cite{fruit_push} to grasp ripe fruit while pushing past unripe ones. In addition, pruning and fruit manipulation have been jointly demonstrated in \cite{bimanual}, where one arm grasps the fruit, and the other prunes it. This work also regulates the force of the manipulators to prevent damaging the fruit \cite{bimanual}. In addition to manipulation, the Push Past Greens methodology \cite{push_greens} reduces occlusions by employing a self-supervised technique to build a policy for moving plant foliage. However, this approach focuses solely on clearing obstructive branches from view rather than manipulating them to specific target locations or considering the delicacy of plant interactions. 

A framework for whole-plant contact interaction was also presented based on a graph neural network, which predicts the plant's state after force application \cite{tree_manipulation}. However, these results are limited to simulations using simplistic tree models with rigid links and PD controllers that may not accurately reflect real tree dynamics. A more advanced simulator incorporating branch dynamics was introduced by learning from real-time probing of actual trees \cite{branch_dyanmics}. While this approach is a significant step forward, its reliance on real-time probing limits its generalization across different plants.

To address these limitations, this work aims to develop a more generalizable approach to branch manipulation, incorporating real-time branch feedback to enable safe and adaptive re-planning. This strategy seeks to balance accuracy and generalizability, allowing robotic systems to handle various branches effectively.

\begin{figure*}[h!]
    \centering
    \includegraphics[trim=1cm 1.5cm 1cm 5cm,clip,width=0.99\textwidth]{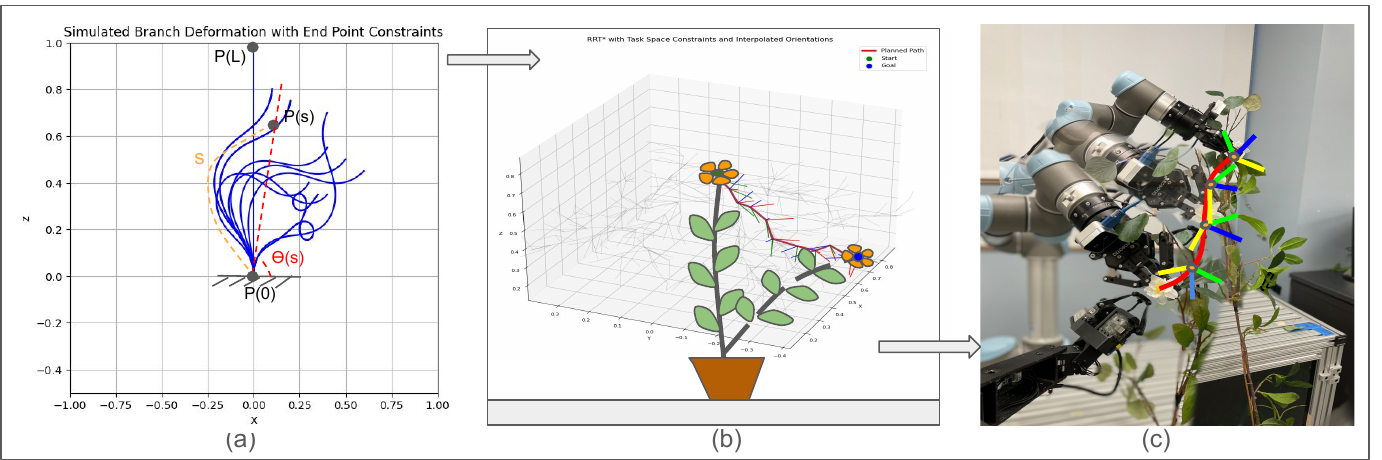}
    \caption{(a) Geometric modeling of deformable linear objects (DLO) provides the general idea which branch configuration are safe to manipulate and which ones are prone to break the branches. (b) Guided by the deformable model, RRT* can sample configurations efficiently to generate the path to the goal region so that the path is safe to manipulate the branch. (c) With the help of force feedback, the path is re-planned to another goal point that is within the field of view of a camera attached to a black robot arm such that the force remains below a certain threshold to prevent branches from getting damaged}
    \label{fig:overall_pipeline}
\end{figure*}

\section{Problem Statement}
Branches are natural objects with varying degrees of flexibility, structural strength, and potential for damage when mishandled. To model such behavior, we assume the branch's shape to be a continuous curve parametrized by \(\mathbf{P}(s)\), where \(s\) is the distance along the branch, as shown in Fig. \ref{fig:overall_pipeline}.a. The base of the branch \(\mathbf{P}(0)\) is fixed. The total length of the branch from its base to the grasp point is \(\mathbf{L}\), which remains constant during the task. It is assumed that the grasping point \(\mathbf{P}(\mathbf{L})\) is predetermined in the branch and serves as the start point \(S\) which is defined by a position \(\mathbf{X} \in \mathbf{R}^3\) and orientation \(\mathbf{O} \in \textbf{so}(3)
\)  of the end effector that grasps the branch. Goal configuration \(G\) lies in the range of the acceptable region defined by radius \(R\) within the field of view of another arm, as shown in Fig. \ref{fig:intro}.c. When manipulating a branch from one configuration to another, the robot must consider the branch's constraints and deformable properties to avoid causing breakage.  Hence, our objective is to generate way-points \(\boldsymbol{\zeta}\) from \(S\) to \(G\) in the constrained workspace \(\mathbf{C}_{ws}\) that can be reached by exerting force \(\mathbf{F}\), which are below the safe threshold \(\mathbf{F}_{\text{threshold}}\) so that branches are not damaged during manipulation.

\section{Technical Approach}

This section describes the proposed approach to obtain a branch model that aids way-point generation and satisfies branch deformation constraints, along with an online force feedback method to manipulate branches safely.

\subsection{Geometric Modeling}
This work utilizes Geometric modeling to estimate the trajectory of branches under manipulation. This model, however, provides only approximate estimates of the deformation. Geometric modeling is generalizable to any branch and computationally efficient. The model is based on differential geometry, where constrained optimization solves the deformable objects' shape. The model is simplified to 2D for the branch manipulation task to aid real-time constraints. The total potential energy of the branch is obtained from \cite{Wakamatsu2004} and calculated by \eqref{potential_energy}:

\begin{equation}
U_{\text{potential}} = \frac{1}{2} \int_{0}^{L} EI \left( \frac{d\theta(s)}{ds} \right)^2 ds
\label{potential_energy}
\end{equation}

where  \( E \) represents young's modulus, \(I\) represents moment of intertia  and $\theta(s)$ represents the orientation of any point $P(s)$. Here, \( \theta(s) \) can be expressed as a linear combination of four kinds of basis functions \( e(s) \) given by:$e_1 = 1$, $e_2 = s$, $ e_{2i+1} = \sin \left(\frac{2\pi i s}{L} \right)$ and $ e_{2i+2} = \cos \left(\frac{2\pi i s}{L} \right)$ which are similar to that of Fourier series basis in representation of any parametric curve.

The endpoint constraints of the branch are defined by the equality constraints as follows:
\begin{align}
    L &= \int_{0}^{L} \left( \cos(\theta(s)) + \sin(\theta(s)) \right) ds \\
    P(L) &= (x_{\text{desired}}, y_{\text{desired}}) \\
    \theta(0) &= \pi/2 \\
    \theta(L) &= \theta_2
\end{align}

Finally, the minimum energy curve is found to satisfy these equality constraints.

\begin{equation}
\label{eq:min_potenital}
\min_{\mathbf{a}} \frac{1}{2} \int_{0}^{L} EI \left( \frac{d\theta}{ds} \right)^2 ds
\end{equation}

\subsection{Safety Zones}
\begin{figure}[h!]
    \centering
    \includegraphics[trim=1cm 0cm 5cm 3cm,clip,width=0.35\textwidth]{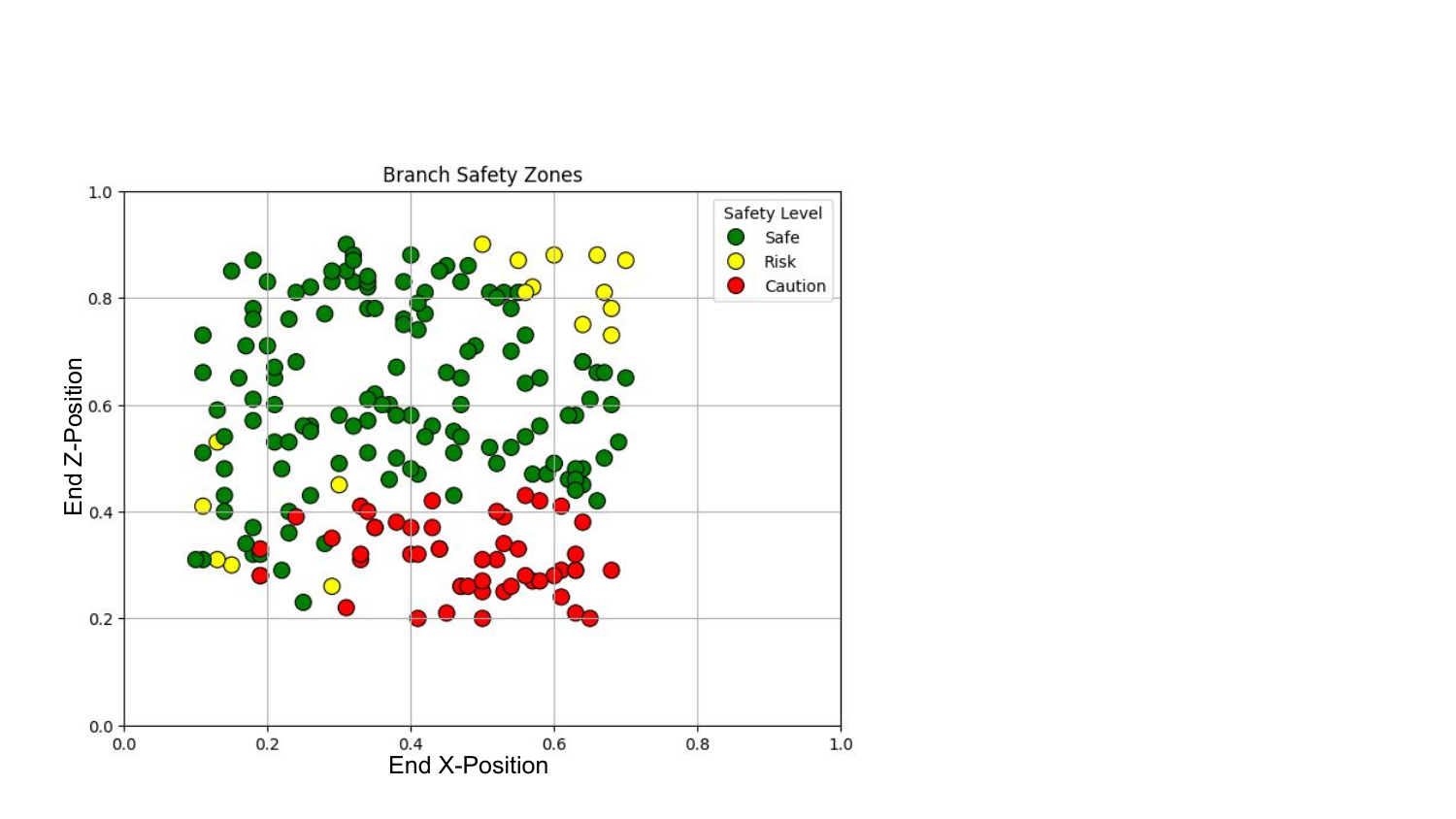}
    \caption{Geometric modeling provides knowledge of any random endpoint constraints' manipulation feasibility and safety level.}
    \label{fig:safety_regions}
\end{figure}

 To solve the constrained optimization problem defined by equation (\ref{eq:min_potenital}), an open source software CasADi\cite{CasADI} is used. To generate the map of the safety region, 200 end-points in an X-Z plane, with the origin being the start, were randomly sampled. Those reachable points that allow smooth branch manipulation are classified as safe points. Those points where the branch intermediate z-value goes above the end-point z-value are classified as caution points. Finally, those points that lead the branch configuration to loop and whose configuration cannot be solved within the constant branch length of $L$ are classified as risky points. The obtained map that shows different safety zones and the simulated branch configuration are shown in Fig.\ref{fig:safety_regions} and Fig.\ref{fig:overall_pipeline}.a, respectively.

\subsection{Constrained Way-points Generation}

Kino-dynamic motion planning can be done with sampling-based motion planning that utilizes a tree-based structure. The RRT* algorithm\cite{karaman2011sampling} is used due to its rewiring step, which refines the path after each iteration. This basis algorithm is modified to obtain a path that satisfies our constraints, as shown in bold in Algorithm \ref{alg:1}. The algorithm randomly samples $q\_{rand}$ in the robot workspace and eliminates samples that violate the branch constraints. The first geometric constraint that needs to be satisfied is that the sample shall not lie beyond a hemisphere defined by the height of the branch.

\begin{equation}
  \label{eq:kinematic_branch_constrain}
  \left( p_{ex}-q_{randx} \right)^2 +\left( p_{ey}-q_{randy} \right)^2+\left( p_{ez}-q_{randz} \right)^2\leq(h)^2
\end{equation}

Where $p=[p_{ex},p_{ey},p_{ez}]$ is the grasp point of the branch, $q_{rand}=[q_{randx},q_{randy},q_{randz}$] is the sampled point and $h$ is the height of the branch at the grasp point. Nearest neighbors are chosen based on the radius $r$ within distance $\delta$. The algorithm steers towards this random point using a line vector ($\Vec{q}_{rand}-\Vec{q}_{near}$) with step $\Delta q$ to obtain $q_{new}$. $q_{new}$ is then verified to be safe and, if so, added as a node in the tree. 

Since branches are delicate, they cannot safely move to all the configurations defined by the geometric constraint  (\ref{eq:kinematic_branch_constrain}). The branch heuristic model is utilized to constrain the growth of RRT*. The safety status of the  $q_{new}$ is then checked using the geometric model and is eliminated if it falls in the risky and caution zones. 

These general heuristics help to minimize re-planning attempts. The node cost is the distance between parent and child $C_{dist}$ plus a penalty $c_{penalty}$ which is determined at runtime only when re-planning of path ($\zeta$) is needed due to dynamic infeasibility of the path which will be described in next section. 

\begin{algorithm}
\caption{RRT* with Task Space Constraints}
\label{alg:1}
\begin{algorithmic}[1]
\Require Start position $q_{start}$, Goal Region $G$, Plant constraints obtained from geometric modeling $q_{safe}$, Maximum iterations $N$, Step size $\Delta q$
\Ensure Constrained path from $q_{start}$ to $G$

\State Initialize tree $T$ with $X$
\For{$i = 1$ to $N$}
    \State\textbf{ Sample $q_{\text{rand}}$ obeying branch constraints}
    \State\text{ $q_{\text{near}} \gets$ Nearest Neighbor($T, q_{\text{rand}}$)}
    \State $q_{\text{new}} \gets$ Steer($q_{\text{near}}, q_{\text{rand}}, \Delta q$)
    \If{\textbf{Geometric\_Model($q_{new}$) == \text{safe}} }
        \State\textbf{set\_node\_cost$(q_{new},\zeta)$}
        \State Parent $q_{\text{min}}$ among neighbors
        \State Add $q_{\text{new}}$ to $T$ with $q_{\text{min}}$ as parent
        \State\textbf{ Rewire $q_{\text{neighbor}}$}
    \EndIf
\EndFor

\State \Return $\zeta$=Path($T,G$)

\end{algorithmic}
\end{algorithm}

Points are sampled in the robot's task space as the kinematics constraints are easier to describe than in the robot's joint space. Though the position can be randomly sampled, doing so for Euler angles will cause jagged robot motions. Hence, unit quaternions instead of Euler angles are utilized for the end-effector pose at the start and final configuration. Then, the intermediate orientations are interpolated to generate smooth motion since the end effector's start and end poses are known. Spherical Linear Interpolation (SLERP)\cite{slerp} is used to interpolate between these two known orientations to ensure smooth and constant-speed rotation. Given two quaternions, $\mathbf{Q}_1$ and $\mathbf{Q}_2$, the interpolated quaternion at time $t \in [0,1]$ is computed as:

\begin{equation}
\mathbf{Q}(t) = \frac{\sin{((1 - t) \gamma)}}{\sin{\gamma}} \mathbf{Q}_1 + \frac{\sin{(t \gamma)}}{\sin{\theta}} \mathbf{Q}_2,
\end{equation}

where $\gamma$ is the angle between $\mathbf{Q}_1$ and $\mathbf{Q}_2$, given by
$\cos{\gamma} = \mathbf{Q}_1 \cdot \mathbf{Q}_2$.

Generated way-points $\zeta$ are followed using pose servo method given by: 
\[
\dot{\mathbf{q}} =  \mathbf{J}^{+}\mathbf{k_{p}} \mathbf{e},
\]
where \( \mathbf{J}^{+} \) is the Jacobian pseudo-inverse, and \( k_{p}\) is the gain and $e$ is the error between current and desired pose.

\subsection{Force Aware Motion Planning }
 To ensure further protection of the branches during manipulation, real-time force feedback is obtained from the force sensor placed on the robot's end effector. While following the way-points $\zeta$, if the force exceeds the threshold $F>F_{threshold}$, re-planning the way-points is performed from that state to move the branch to another configuration within the acceptable goal region $G$. In the re-planing stage, the nodes' rewiring is done so that the new path does not align with the previous path, which caused the robot to exert excessive force beyond the threshold to move the branch; via a penalty term which is added to the cost of the $q_{new}$ nodes within the distance $r'$ to the way-points $\zeta$. 
 
 \begin{equation}
   \mathbf{c_{penalty}}(q_{new'}) = \frac{1}{min(dist(q_{new'},\zeta))+\epsilon}  
 \end{equation}
 
Where $dist()$ calculates the distance between the point $q_{new'}$ and array of way-points $\zeta$, min() returns the  finds the minimum distance to the path, and $\epsilon$ is the small value used to avoid division by zero.

\section{Experiments and Results}
In this section, experimental setup, branch constraint aware planning, and performance of the force-aware planning approach are presented.

\begin{figure*}[h]
     \centering
	 \begin{subfigure}{0.28\textwidth}
	  \includegraphics[width=\linewidth]{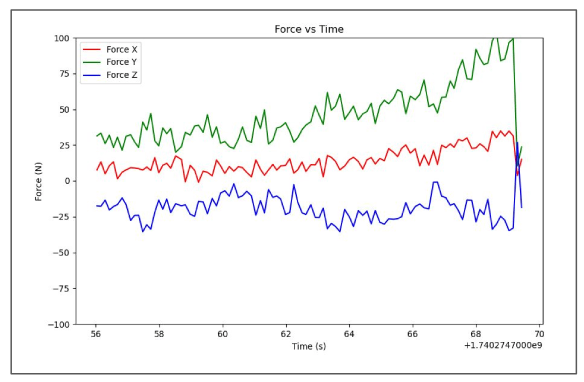}
		\caption{ }
        \label{fig:direct_force}
	 \end{subfigure}
       \begin{subfigure}{0.35\textwidth}
        \includegraphics[width=\linewidth] {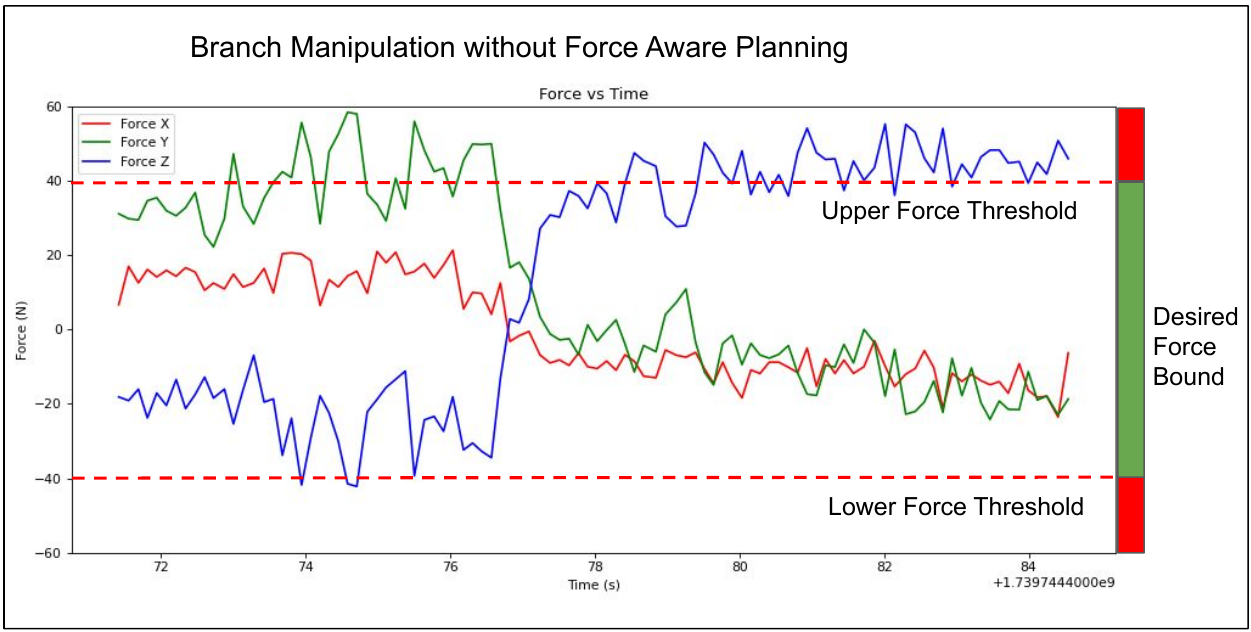}
 	 \caption{ }
 	 \label{fig:force_unaware}
 	 \end{subfigure}
       \begin{subfigure}{0.29\textwidth}
 	  \includegraphics[width=\linewidth]{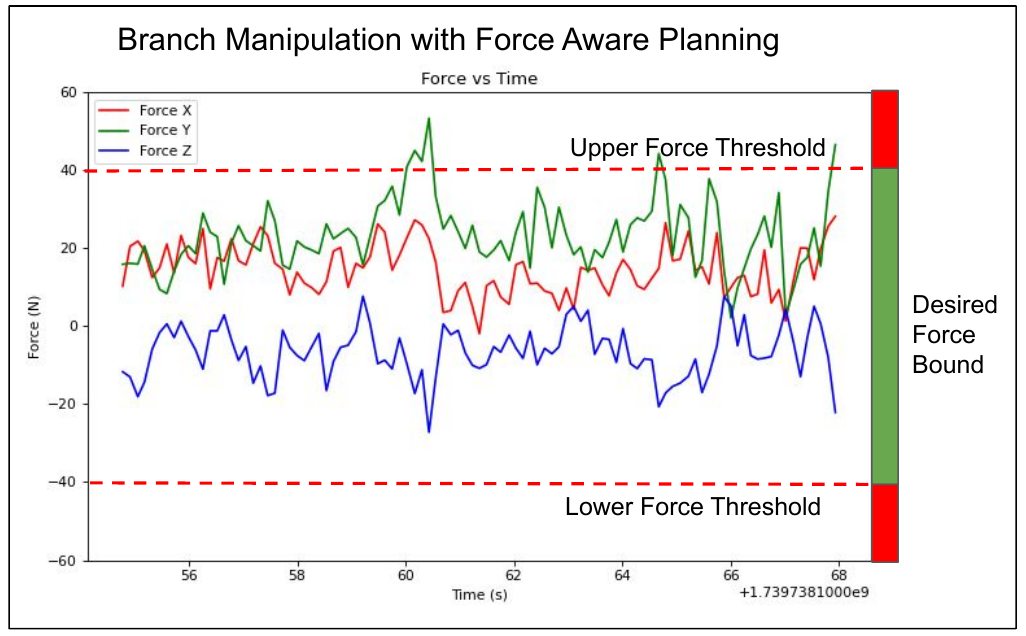}
 		\caption{}
 		\label{fig:force_aware}
 	    \end{subfigure}
        
    \caption{(a) When moving a branch without branch model information or force feedback, significant manipulation force (100 N) occurs. (b) Introducing branch model information reduces the manipulation force to below 60 N (c). Utilizing both branch information and force feedback maintains the applied force below the 40 N Force Threshold.}
    \label{fig:boundedforce}
 \end{figure*}

\subsection{Experimental Setup}
The branch manipulation experiment was conducted using a single vertical artificial branch with artificial flowers and leaves. The task involved grasping the branch and orienting a flower to face a camera mounted on a separate robotic arm designed for pollination \cite{smith2024design}. The primary manipulating robot was a UR5 arm equipped with a Robotiq FT300 force sensor at its end effector, capable of measuring forces up to 300 N along all three axes. Fig. \ref{fig:overall_pipeline}.c illustrates the experimental setup and the intended task.

\subsection{Branch Constraint Aware Planning}

This experiment compares the proposed algorithm without force-aware replanning to the baseline RRT* for manipulating the artificial branch. Figure \ref{fig:direct_force} shows that baseline RRT* often resulted in excessive force, exceeding 100 N, causing the branch to detach and the force to drop to zero. In contrast, incorporating branch geometry and safety zones into the proposed algorithm limited the force to below 60 N (Figure \ref{fig:force_unaware}) by guiding the branch through waypoints that minimize internal potential energy without violating constraints.

The reduced force is due to the branch-informed planner's ability to avoid caution zones, as shown in Figure \ref{fig:safety_regions}, which are unknown to baseline RRT*. The intermediate waypoints and orientations generated by Algorithm \ref{alg:1} are displayed in Figure \ref{fig:path_with_orient}.

 \subsection{Force Aware Planning}
\begin{figure}[h]
      \centering
	   \begin{subfigure}{0.48\linewidth}
		\includegraphics[trim=1cm 1cm 1cm 0cm,clip,width=1\textwidth]{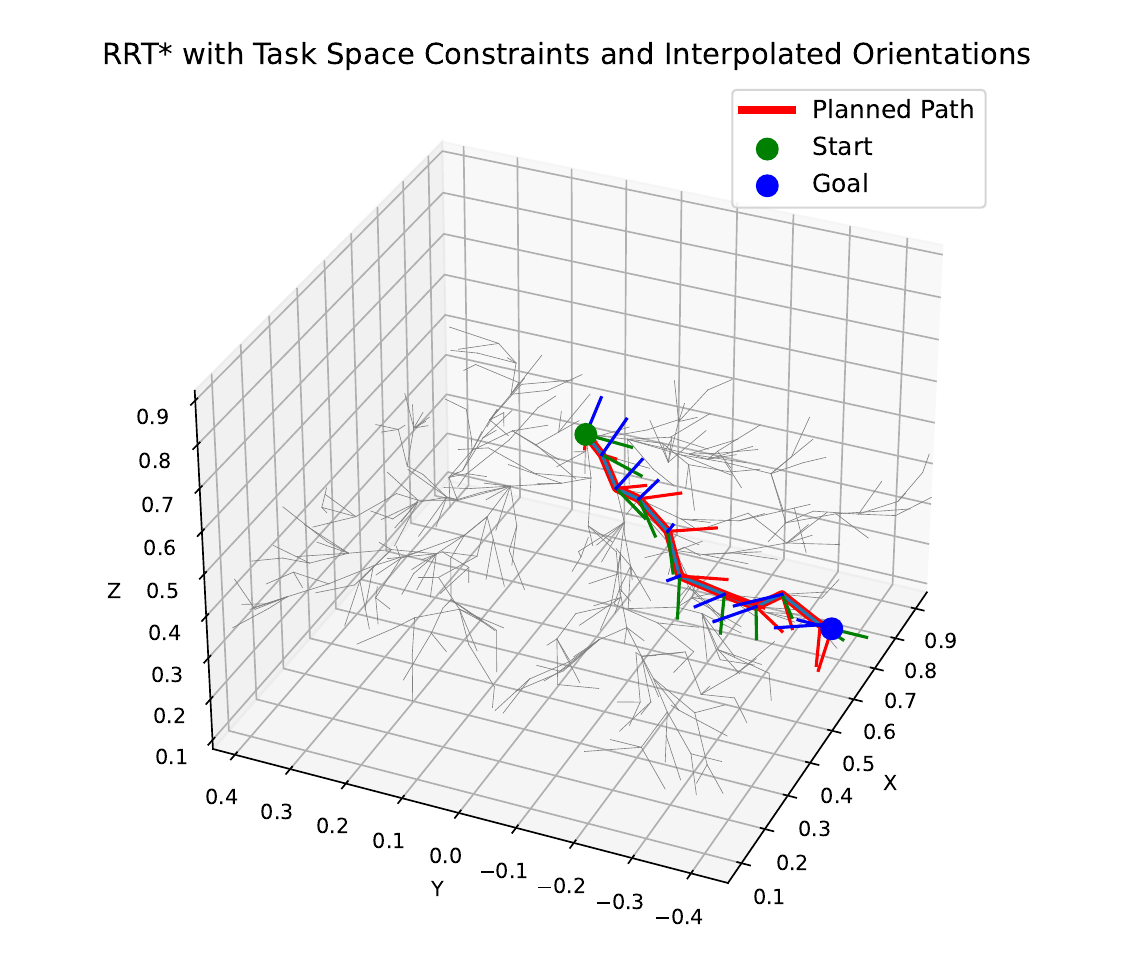}
		\caption{}
		\label{fig:path_with_orient}
	   \end{subfigure}
          \begin{subfigure}{0.48\linewidth}
		 \includegraphics[trim=21cm 6cm 18cm 8cm,clip,width=\textwidth]{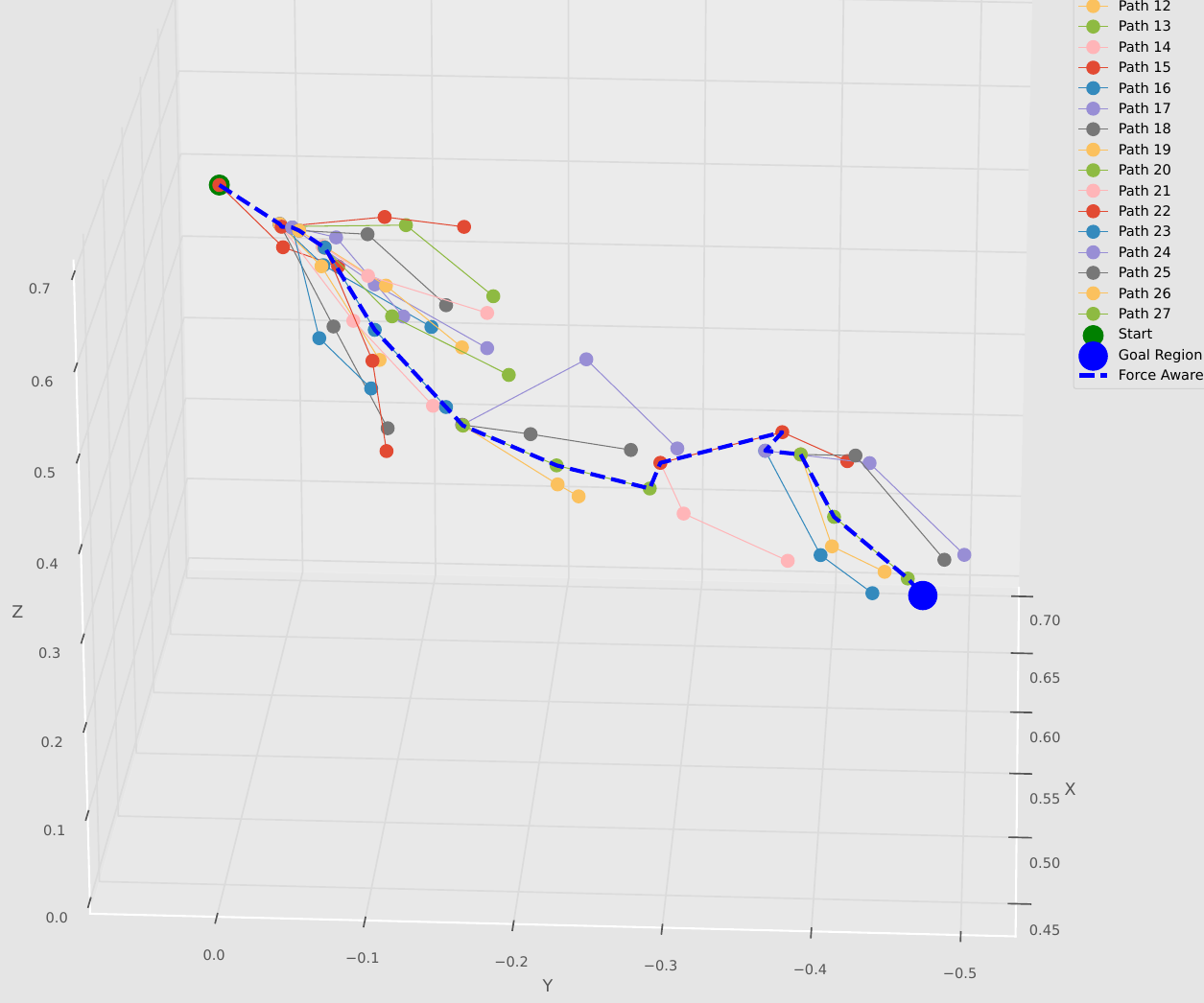}
		 \caption{}
		 \label{fig:repalned_path}
	      \end{subfigure}
	 \caption{(a) Path with orientation to manipulate branch from start configuration to goal region. (b) Among the different planned path to reach the goal, the planner followed the dotted line which minimized the force exerted on branch. }
	\label{}
\end{figure}

This experiment evaluates the impact of adding online force-aware re-planning to the proposed branch-informed planning algorithm. The manipulation force with force-aware re-planning is shown in Figure \ref{fig:force_aware}, while the planned and re-planned paths are displayed in Figure \ref{fig:repalned_path}. The experiment involved five different starting poses targeting a common goal location, with each configuration repeated 10 times for a total of 50 trials. A plan was successful if it brought the grasp point within a 5 cm radius of the goal point. The planning time limit is 400 seconds, and the force bound is between -40N and 40N.

The first observation from Fig. \ref{fig:force_aware} is that online force-aware re-planning reduced the manipulation force from 60 N to below the desired force threshold of 40 N. Thus, the online re-planner can successfully alter the path to reduce the damage to the plant. Table \ref{table:performance}, displays that out of the 50 total attempts, there were 39 successful attempts and 11 failure cases. The average number of re-planning attempts across all scenarios is 20, as mentioned in Table \ref{table:performance}, and an example of a re-planned path is shown in Fig. \ref{fig:repalned_path}.

\begin{table}[h!]
\renewcommand{\arraystretch}{1.2}
\Huge
\centering 
\caption{Performance of the planner evaluated using five different start poses and a common goal point in the field of view, each repeated 10 times.}
\resizebox{0.48\textwidth}{!}{
\begin{tabular}{|l|l|l|l|l|l|l|l|}
\hline
\textbf{\begin{tabular}[c]{@{}l@{}}Common \\ Goal\end{tabular}} & \textbf{[0.45,-0.43,0.36]} & \textbf{\begin{tabular}[c]{@{}l@{}}Euclidean\\ Distance(m)\end{tabular}} & \textbf{\begin{tabular}[c]{@{}l@{}}Path\\ Length(m)\end{tabular}} & \textbf{\begin{tabular}[c]{@{}l@{}}Goal \\ offset(m)\end{tabular}} & \textbf{\begin{tabular}[c]{@{}l@{}}Planning\\ Time(s)\end{tabular}} & \textbf{\begin{tabular}[c]{@{}l@{}}Replanning \\ Attempts\end{tabular}} & \textbf{Sucess} \\ \hline
Start-1                                                         & {[}0.66,0.046,0.52{]} & 0.544                                                           & 0.829                                                    & 0.018                                                     & 183 (± 9.44)                                                      & 16 (± 2.08)                                                             & 8      \\ \hline
Start-2                                                         & {[}0.65,0.07,0.62{]}  & 0.598                                                           & 0.921                                                    & 0.026                                                     & 201 (± 12.01)                                                       & 19 (± 2.31)                                                           & 8      \\ \hline
Start-3                                                         & {[}0.81,-0.007,0.4{]} & 0.557                                                           & 0.743                                                    & 0.017                                                     & 176 (± 11.87)                                                       & 16 (± 2.22)                                                            & 8      \\ \hline
Start-4                                                         & {[}0.63,0.12,0.58{]}  & 0.619                                                           & 1.126                                                    & 0.023                                                     & 221 (± 12.59)                                                       & 25 (± 3.61)                                                            & 7      \\ \hline
Start-5                                                         & {[}0.62,0.15,0.43{]}  & 0.608                                                           & 1.064                                                    & 0.019                                                     & 208 (± 10.52)                                                       & 23 (± 3.35)                                                            & 8      \\ \hline
\end{tabular}
}
\label{table:performance}
\end{table}

As shown in Table \ref{table:performance}, an increase in the standard deviation of re-planning attempts and a decrease in the success rate were observed when the Euclidean distance increased by adjusting the Y-position, as seen in start-3 and start-4. This is likely due to the branch being modeled only in the X-Z plane. Implementing a more accurate 3D model, despite higher computational costs, could potentially mitigate this issue. Failure cases also occurred when the pose servo needed to rotate the branch significantly between interpolations to reach the target orientation. These failures increased at shorter Euclidean distances, where fewer interpolated waypoints resulted in larger orientation adjustments.

\section{Conclusion and Future Work}
This paper developed a method of manipulating branches to assist different agricultural tasks by utilizing a geometric plant model to guide the RRT* to generate safe paths and re-plan the path using online force feedback if the force goes beyond the desired threshold. The experimental result shows that the proposed planner can bound the force to the desired limit so that the robot does not damage the branches by exerting excessive force. Unlike previous methods, which rely on Finite element methods (FEM) and mass spring model of the branches, our model does not need individual branch parameters or interactions for tuning. Hence, this method can generalize well in manipulating various tree branches as it can handle uncertainties by re-planning motion based on online force feedback during real-time manipulation. The limitation of this method lies in determining the safe force threshold, as different types of branches require different force thresholds for safe manipulation. Future work in this direction is to learn the safe force threshold based on the geometry of the branch, estimate the suitable grasp point, and design a compliant gripper with a force sensor suitable for manipulating branches. 

\section*{Acknowledgment}
We extend our gratitude to Dr. Nicole Waterland and her students for providing access to the greenhouse facility and plants.

\bibliographystyle{IEEEtran}

\bibliography{references}

\addtolength{\textheight}{-12cm}   

\end{document}